\documentclass[runningheads]{llncs}
\usepackage{cite}
\usepackage{graphicx}
\begin{document}
%
%\title{Deception Detection with Feature-Augmentation by soft Domain Transfer\thanks{Supported by organization x.}}
\title{Deception Detection with Feature-Augmentation by soft Domain Transfer}


\authorrunning{Shahriar et al.}
% \author{Sadat Shahriar\inst{1}\orcidID{0000-1111-2222-3333} \and
% Arjun Mukherjee\inst{2,3}\orcidID{1111-2222-3333-4444} \and
% Omprakash Gnawali\inst{3}\orcidID{2222--3333-4444-5555}}
% %
% % \authorrunning{F. Author et al.}
% \institute{University of Houston, Tx 77054, USA \\
% \email{shahria@cougarnet.uh.edu}\\
% \and
% University of Houston, Tx 77054, USA\\
\author{Sadat Shahriar\thanks{Corresponding Author}
 \and Arjun Mukherjee \and Omprakash Gnawali}

\institute{ University of Houston, Houston, Tx 77004, USA \\
\email{sshahria@cougarnet.uh.edu, arjun@cs.uh.edu, odgnawal@central.uh.edu}} 
% typeset the header of the contribution
%

\maketitle
\begin{abstract}
In this era of information explosion, deceivers use different domains or mediums of information to exploit the users, such as News, Emails, and Tweets. Although numerous research has been done to detect deception in all these domains, information shortage in a new event necessitates these domains to associate with each other to battle deception. To form this association, we propose a feature augmentation method by harnessing the intermediate layer representation of neural models. Our approaches provide an improvement over the self-domain baseline models by up to 6.60\%. We find Tweets to be the most helpful information provider for Fake News and Phishing Email detection, whereas News helps most in Tweet Rumour detection. Our analysis provides a useful insight for domain knowledge transfer which can help build a stronger deception detection system than the existing literature. 

\end{abstract}
\section{Introduction}
Deception in the text implies a deliberate attempt of a sender to misconstrue an affair or create a false impression \cite{Burgoon1994}. Deception in the text can occur in multiple domains like News, Tweets, Emails, and research has been done to detect deception in domain-specific settings \cite{shu2017fake, varshney2016survey, zubiaga2018detection}. Although deceivers use their con in each domain with a unique style, all kinds of deception have the same agenda of deceiving people. Hence, detecting deception in one domain can be leveraged with detection in the other domain. In this paper, we perform a soft domain transfer by investigating how to harness the power of deception detection in domain A to detect deception in domain B. We also investigate the effectiveness of domain transfer when the source domain is non-deceptive. 

% Unlike human intelligence, AI models are often prone to start from scratch, having no experience at all and aimed to learn from an input dataset only \cite{goodfellow_article}. However, transfer learning has been a widely researched topic in machine learning \cite{5288526}. Pretrained language model like BERT, ELMo, trained on a wide variety of textual content achieved some level of transfer learning capability, but they are more generic in nature, since they are trained in generic texts like wikipedia and books, and fine-tuned for specific tasks \cite{Devlin2019BERTPO, elmo}. However, for a complex task of detecting deception, a generic fine-tuned model may not be adequate.

Researchers looked at deception from a holistic point of view in the hope of capturing the nuances in the style of deception \cite{shahriar2021domain}. However, it is not clear if such a clear pattern exists since deceptions in different domains are very different. Additionally, further investigation is needed to quantify the ``help'' received from one domain to the other. To this end, there is a significant research gap in achieving the domain knowledge transfer. We define \textit{Deceptive Domain} as different sources of the information through which deception occurs, and we use fake news, phishing emails, and rumours as deception in different domains. As non-deceptive domains, we use Newsgroup topics, sentiment detection, and Wikipedia ontology detection. Therefore, we formulate our first research question as (\textbf{RQ1}): Can knowledge transfer from different domains help improve deception detection? Additionally, our second research question is posed as (\textbf{RQ2}) Between the deceptive and non-deceptive domains, which set of domains are most helpful in detecting deception? 
%Which domain information can be most helpful while detecting deception in another domain? 
To answer these questions, we train six different BERT and LSTM models \cite{Devlin2019BERTPO, LSTM} for three deceptive and three non-deceptive domains. We collect the intermediate-layer information of the target domain and harness the power of the external domain by combining the intermediate-layer and train a Fully-Connected Deep Neural Network (FC-DNN) to detect deception. In this way of feature augmentation, we leverage the knowledge of other domains in the FC-DNN model by injecting that knowledge into the input information.  

The significance of this study is manifold. First, in many domains, deceptive data is significantly scarce. For example, individuals and corporations are reluctant to share the phishing emails they receive to evade embarrassment \cite{ELAetAl:IWSPA-AP2018}. Second, with the influx of social media, the information is flown through different domains when a new event emerges. For example, the emergence of COVID-19 created a significant misinformation upsurge in news, tweets, and Facebook posts; thus, learning deception by relying on one domain only results in missing other domain information. Finally, this study can guide researchers to lay out a selective knowledge transfer scheme from different domains and find the generalized pattern of deception. The novelty of our work is, to the best of our knowledge, this is the first work that explores the effectiveness of domain transfer in deception detection, and opens up new avenues of further promising research.     

\vspace{-.2 cm}
\section{Related Works}

% Cross-Domain deception detection exists in the literature but most of these works defined ``Domain'' as same medium of deception on different topics. Since we focus on different mediums of deception than topics, these definitions are different from our definition.

Hern\'andez-Casta\~neda et al. proposed a cross-domain deception detection using SVM, where the datasets were of opinions on different topics \cite{Hernandez2017}.   They aimed to find a general set of features in different experimental train-test settings. Similar research was done by \cite{SANCHEZJUNQUERA2020122} and \cite{li-etal-2014-towards}.
% and   S\`anchez-Junquera et al. proposed a similar approach using Na\`{i}ve Bayes classifier \cite{SANCHEZJUNQUERA2020122}. Li et al. proposed a generalized approach to detect opinion spam in Hotel, Restaurant and Doctor reviews using a Sparse Additive Generative Model and SVM \cite{li-etal-2014-towards}.
In the Fake News detection task, P\`erez-Rosas et al. performed cross-domain experiment on two different datasets and showed the challenges of generalizability \cite{PrezRosas2018AutomaticDO}. Gautam and Jerripothula used Spinbot, Grammarly and GloVe-based method to for cross-domain fake news detection \cite{Gautam2020SGGSG}.
%Saikh et al. proposed an ELMo and BiGRU-based method for similar research and reported a significant drop in performance from the in-domain experiments \cite{Saikh2019ADL} 
However, these research were done to make the deception detection system topic-agnostic rather than mediums-of-deception agnostic. Hence, harnessing the deception-detection capability in the cross-domain setting remains an unexplored area.

 \section{Dataset}
 
 For the Email domain, we use a phishing email dataset from the Anti-Phishing Pilot at ACM IWSPA 2018 \cite{IWSPA_DATASET}. The training set has 5092 legitimate and 629 phishing emails, and the test data size is 4300, with 3825 legitimate and 475 phishing emails. We label phishing emails as deceptive and non-phishing as non-deceptive. For the News domain, we use LIAR dataset \cite{wang-2017-liar}, which comes with six labels of the news, namely, True, Mostly-True, Half-True, Mostly-False, False, Pants-on-Fire False. We consider the first two as non-deceptive text, and the last four as deceptive following the work in \cite{upadhayay2020sentimental}. 
% In the training set, the Fake and non-Fake news distribution are 792 and 5092, respectively, and in the test set, we have 475 and 3825, respectively. 
 For the Tweet domain, the PHEME dataset is used, which had 2402 rumour texts, and 4023 non-rumour texts \cite{Zubiaga2016}. We label rumour tweets as deceptive and non-rumour tweets as non-deceptive.  
 
For non-deceptive tasks, three datasets are used. The IMDB movie review dataset comes with 50,000 reviews, labeled as positive or negative \cite{maas-EtAl:2011:ACL-HLT2011}. The 20 newsgroups dataset consists of around 18000 samples with labels on newsgroups posts about 20 topics \cite{Lang95}. The Wikipedia topic classification dataset consists of 342,782 articles with 9 topic classes \cite{Lehmann2015DBpediaA}. We randomly sample 10,000 texts for each non-deceptive domain and use 80-20 ratio for the train-test. 

\vspace{-.2 cm}
\section{Methodology}

We use two neural models-- the Bidirectional Encoder Representations from Transformers (BERT) model and Long Short-Term Memory (LSTM) model as baseline methods \cite{Devlin2019BERTPO, LSTM}. The BERT model is
built with transformer layers consisting of encoders and
decoders with self-attention capability. We fine-tune our baseline self-domain BERT model, extract the model's last [CLS] layer, and use an FC layer
and a softmax for the downstream classification
task. LSTMs are an efficient variation of Recurrent Neural
Network (RNN) with added long-term dependency solution. We use the sequence of words as the input of a two-layer LSTM model and use an FC layer on top to classify the text. 

\begin{figure*}
\includegraphics[height=6cm, width=11cm]{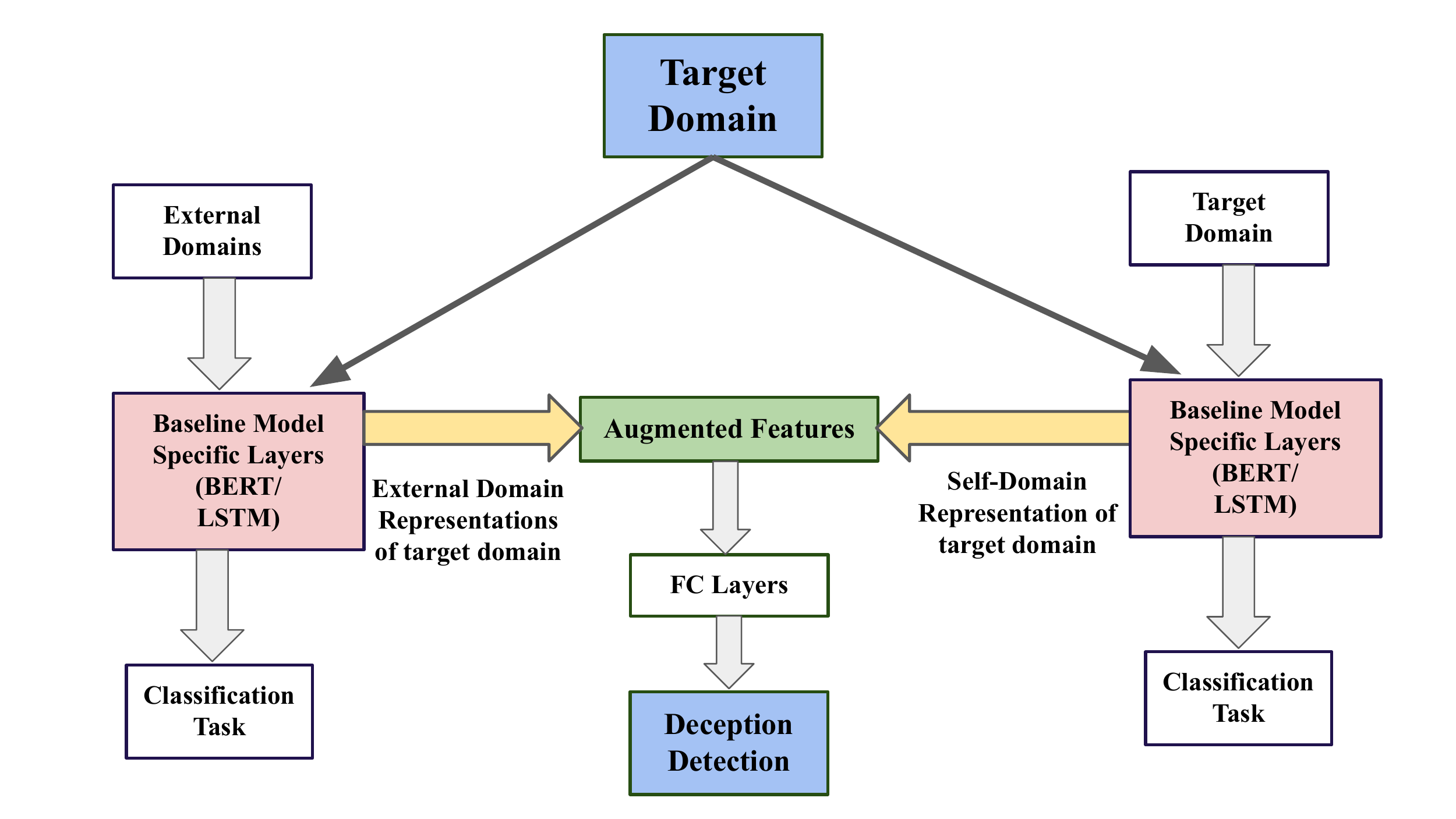}
\caption{Feature augmentation by soft domain transfer to improve deception detection. We augment the deceptive features by concatenating the intermediate layer representation of baseline models of both target and external domains and the augmented features are fed to a FC network to detect deception. } \label{fig:overall}
\end{figure*}

For the feature augmentation process, we perform the Intermediate Layer Concatenation (\textbf{ILC}), which is explained in Figure \ref{fig:overall}. For the BERT model, we extract the self-domain trained [CLS] layers of different domains and concatenate them, representing our target domain's augmented feature set. Similarly, for the LSTM model, we extract the output from the final LSTM layer representation of different domains and concatenate them. The augmented feature set is then fed to a 2-layer Fully-Connected (FC) model to detect deception. Finally, all the network hyperparameters are set using validation sets generated by sampling 20\% data from the training set. 

\section{Results and Discussion}

The Table \ref{tab:results:BERT} shows the performance of feature augmentation with different deception domains using the BERT-based ILC models. For the Email domain, we observe that the News domain improves the F1 score of phishing detection by 2.31\% and the Tweet domain improves the performance by 4.89\%. While both Tweet and News domains are combined, we observe a performance boost in F1-score by 6.60\%. For News and Tweet domain, we also observe an improved performance with deceptive feature augmentation. Emails help detect fake news by 0.75\%, and Tweets help by 1.69\%. Tweet rumour detection gets performance improvement of 1.36\% from News and 1.14\% from Email domain. However, compared to the News and Tweet, performance improvement is higher in the Email domain. Being a  pretrained model, BERT is more likely to perform well with public texts like News and Tweet, and thus the baseline model achieves a better understanding of deception in these two domains. Hence, the augmentation from other deceptive domains improves phishing email detection more than deception detection in other domains.

\begin{table*}
\caption{Cross-Domain deception detection based on BERT models.  \textbf{E}, \textbf{T}, and \textbf{N} stands for Email, Tweet and News respectively. For example, ``ILC-TN'' stands for ILC model where Tweet and News domains are combined. }
\begin{center}
\begin{tabular}{c|cc||cc|cc|cc|cc|}

\cline{2-11} &
\multicolumn{2}{|c||}{\textbf{Baseline -- BERT}} & \multicolumn{2}{|c|}{\textbf{ILC -- EN}} &
\multicolumn{2}{|c|}{\textbf{ ILC -- TN }} &
\multicolumn{2}{|c|}{\textbf{ILC -- ET}} &
\multicolumn{2}{|c||}{\textbf{ILC -- ETN}}\\
\hline \textbf{Domains} & \textbf{F1} & \textbf{ACC}& \textbf{F1} & \textbf{ACC} & \textbf{F1} & \textbf{ACC} & \textbf{F1} & \textbf{ACC} & \textbf{F1} & \textbf{ACC}  \\ \hline
Email & 80.99 & 95.41 & 83.31 & 96.03 & -- & -- & 85.88 & 96.79 & \textbf{87.59} & \textbf{97.39} \\
News & 76.88 & 63.55 & 77.63 & 63.93 & 78.57 & \textbf{67.80} & -- & -- & \textbf{78.80} & {67.56}\\
Tweet & 80.34 & 84.79  & -- & -- & 81.70 & 86.23 & 81.48 & 85.99 & \textbf{82.07} & \textbf{86.77}   \\
\hline
\end{tabular}
\end{center}
\label{tab:results:BERT}
\end{table*}

\begin{figure*}
\includegraphics[height=3cm, width=11cm]{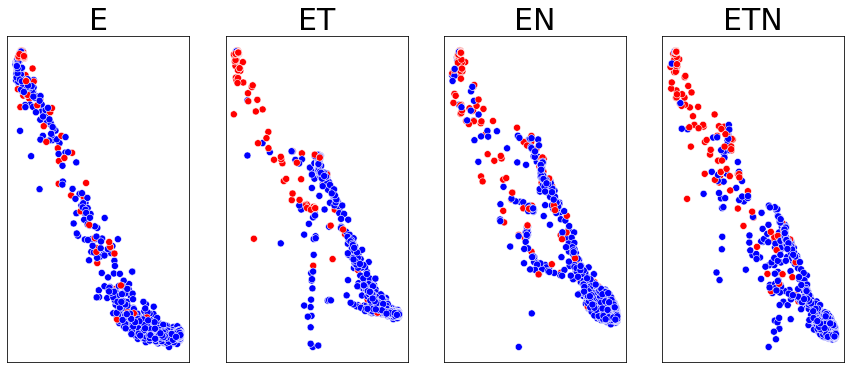}
\caption{SVD-reduced representation (BERT model) of Email domain with their self-domain features vs intermediate layer concatenated features with different deception domains. Blue points represent non-deception and red points represent deception } \label{fig:SVD}
\end{figure*}

We further investigate the effectiveness of cross-domain
feature augmentation by projecting the data to a 2-D subspace
using Singular Value Decomposition (SVD) method. Figure \ref{fig:SVD} clearly shows an improved feature separation while Tweet and News domains are added with Email, increasing the distance between the deceptive and non-deceptive samples' center of cloud by 50.23\%, when all three domains are concatenated.

Using the LSTM-based feature augmentation technique, we compromise overall performance, but unlike BERT, we do not use a pretrained model. Therefore, we observe a consistent performance improvement in all three deception domains (Table \ref{tab:results:LSTM}). In the Email domain, like the BERT-based ILC model, the Tweet domain helps the most, and overall improvement is up to 3.97\%, with a combined augmentation. Tweets are the most helpful domain both for Email and News. However, the best performance is obtained while all three domains are combined, giving a performance raise of 4.82\% in the News domain and 3.39\% in the Tweet domain. As standalone domains, News helps the Tweet domain most, providing a boost of 1.76\% in the F1-score. 

\begin{table*}
\small
\caption{Cross-Domain deception detection based on LSTM models. }
\begin{center}
\begin{tabular}{c|cc||cc|cc|cc|cc|}

\cline{2-11} &
\multicolumn{2}{|c||}{\textbf{Baseline -- LSTM}} & \multicolumn{2}{|c|}{\textbf{ILC -- EN}} &
\multicolumn{2}{|c|}{\textbf{ ILC -- TN }} &
\multicolumn{2}{|c|}{\textbf{ILC -- ET}} &
\multicolumn{2}{|c||}{\textbf{ILC -- ETN}}\\
\hline \textbf{Domains} & \textbf{F1} & \textbf{ACC}& \textbf{F1} & \textbf{ACC} & \textbf{F1} & \textbf{ACC} & \textbf{F1} & \textbf{ACC} & \textbf{F1} & \textbf{ACC}  \\ \hline
Email & 71.23 & 92.23 & 72.04 & 92.45 & -- & -- & 74.96 & 94.68 & \textbf{75.20} & \textbf{94.87} \\
News & 72.71 & 62.03 & 74.69 & 62.43 & 75.45 & {63.18} & -- & -- & \textbf{77.53} & \textbf{63.59}\\
Tweet & 73.11 & 77.16  & -- & -- & 74.87 & 79.29 & 74.18 & 79.00 & \textbf{76.50} & \textbf{82.13}   \\
\hline
\end{tabular}
\end{center}
\label{tab:results:LSTM}
\end{table*}

\begin{table}
\caption{BERT-based Deception detection by feature augmentation from non-deceptive domains.}
\begin{center}
\begin{tabular}{c|cc|cc|cc|cc|}

\cline{2-9} &
\multicolumn{2}{|c|}{\textbf{Sentiment}} & \multicolumn{2}{|c|}{\textbf{Newsgroup}} &
\multicolumn{2}{|c|}{\textbf{ Wikipedia }} &
\multicolumn{2}{|c|}{\textbf{Combined}}\\
\hline \textbf{Domains} & \textbf{F1} & \textbf{ACC}& \textbf{F1} & \textbf{ACC} & \textbf{F1} & \textbf{ACC} & \textbf{F1} & \textbf{ACC}\\ \hline
Email & 81.24 & 96.09 & 80.95 & 96.14 & 81.04 & 96.13 & 81.26 & 96.09  \\
News & 77.43 & 63.10 & 77.58 & 63.19 & 77.89 & 63.51 & 78.05 & 63.79 \\
Tweet & 81.41 & 86.07  & 81.32 & 85.91 & 81.24 & 85.80 & 81.89 & 86.30  \\
\hline
\end{tabular}
\end{center}
\label{tab:results:Ndecep:BERT}
\end{table}

Next, we investigate deception detection performance while augmented with non-deceptive domains using BERT models. From Table \ref{tab:results:Ndecep:BERT}, we observe that Sentiment and Wikipedia slightly improve the performance of phishing email detection, and with combined domains, it improves by 0.27\% in F1-score. For the News domain, Wikipedia helps the most, and overall we get a 1.16\% improvement in F1 score with all the domains combined. The Sentiment is the most helpful domain for detecting rumour in Tweets, improving the performance by 1.07\% in the F1-score, and with the combined domains, the improvement is  1.55\%. We also find a similar performance with LSTM-based ILC models, with the best performance in combined domains, improving the Email, News, and Tweet domain deception detection by 2.60\%, 5.04\%, and 3.10\% respectively (Table \ref{tab:results:Ndecep:LSTM}). 

From the above discussion, we find that the feature augmentation from different domains helps improve the deception detection task. However, the performance boost is greater when the external domain is deceptive than a non-deceptive one, and thus, a soft domain transfer takes place. 

\begin{table}
\caption{LSTM-based Deception detection by feature augmentation from non-deceptive domains. }
\begin{center}
\begin{tabular}{c|cc|cc|cc|cc|}

\cline{2-9} &
\multicolumn{2}{|c|}{\textbf{Sentiment}} & \multicolumn{2}{|c|}{\textbf{Newsgroup}} &
\multicolumn{2}{|c|}{\textbf{ Wikipedia }} &
\multicolumn{2}{|c|}{\textbf{Combined}}\\
\hline \textbf{Domains} & \textbf{F1} & \textbf{ACC}& \textbf{F1} & \textbf{ACC} & \textbf{F1} & \textbf{ACC} & \textbf{F1} & \textbf{ACC}\\ \hline
Email & 73.18 & 93.41 & 71.49 & 92.18 & 72.17 & 92.77 & 73.83 & 94.11  \\
News & 72.80 & 61.97 & 72.96 & 62.11 & 75.50 & 63.37 & 76.48 & 63.41 \\
  Tweet & 73.93 & 78.32  & 73.00 & 77.91 & 75.17 & 81.56 & 76.21 & 82.02  \\
\hline
\end{tabular}
\end{center}
\label{tab:results:Ndecep:LSTM}
\end{table}
\vspace{-1 cm}

\section{Conclusion and Future Work}
Despite the research on deception detection in many existing domains, there is a research gap on how to harness cross-domain deception detection by transferring the knowledge gained from one domain to the other. In this paper, we bridge the gap using an intermediate-layer concatenation approach from the neural model.
%We found that deceptive domains contribute more to the performance boost than the non-deceptive domains.
%Our research will be beneficial to the domains where adequate data is lacking, and also on the emerge of a new event like COVID-19. 
There are several future research directions for this work. First, our analysis is limited to three domains only. Several other domains, e.g., reviews, Facebook posts, and Whatsapp message forwards, can also be explored for cross-domain deception detection. Furthermore, we use only one dataset in each domain. Additional research with more datasets in these domains will help solidify our hypothesis.

\subsubsection*{Acknowledgements}
The research was supported in part by grants NSF 1838147, ARO W911NF-20-1-0254. The views and conclusions contained in this document are those of the authors and not of the sponsors. The U.S. Government is authorized to reproduce and distribute reprints for Government purposes notwithstanding any copyright notation herein.

\bibliographystyle{splncs04}
\bibliography{mybibliography}

\end{document}